\documentclass{llncs}

\usepackage{makeidx}  

\begin{document}

\title{A study on the relation between linguistics-oriented and domain-specific semantics}

\author{He Tan}

\institute{Institutionen f\"{o}r datavetenskap, \\
Link{\"o}pings universitet, Sweden}

\maketitle

\begin{abstract}

In this paper we dealt with the comparison and linking between lexical resources with domain knowledge provided by ontologies. It is one of the issues for the combination of the Semantic Web Ontologies and Text Mining. We investigated the relations between the linguistics-oriented and domain-specific semantics, by associating the GO biological process concepts to the FrameNet semantic frames. The result shows the gaps between the linguistics-oriented and domain-specific semantics on the classification of events and the grouping of target words. The result provides valuable information for the improvement of domain ontologies supporting for text mining systems. And also, it will result in benefits to language understanding technology. 

\end{abstract}

\section{Introduction}

The combination of Ontologies and Text Mining (TM) has been considered important within the life science research community, to keep track of knowledge reported in the increasing accumulation of scientific literature available online (e.g.~\cite{WKB07}). Although most of them have not been developed for natural language analysis~\cite{TL09}, ontological resources provide domain knowledge required by text mining systems. 
 Ontologies can provide a structured and semantically richer representation for text mining systems. They will provide the capability of natural language processing (NLP) systems to run reasoning over natural language. Furthermore, they will provide the framework for a consistent semantic integration of the various resources used throughout a text mining system, and also the integration of knowledge extracted from text and from other different resources like databases. This is consistent with the vision of the Semantic Web.
 
In this paper we dealt with one of the issues for the combination of the two technologies, comparing and linking lexical resources with domain knowledge provided by ontologies. 
Neither ontologies nor their interplay with the lexical resources have received much attention in the development of lexical resources for the bio-medicine domain, although in very beginning domain ontologies have been considered important resources.  
In this paper we explicitly linked the FrameNet~\cite{fn} semantic frames with the concepts from the biological process ontology of the Gene Ontology (GO)
The result shows the gaps between the linguistics-oriented and domain-specific semantics on the classification of events and the grouping of target words. It provides valuable information for the improvement of domain ontologies support for text mining systems. And also, it will result in benefits to language understanding technology. 

The Gene Ontology
has been widely used as knowledge base for NLP systems in the domain. Among the three subontologies of terms, biological process ontology provides structured knowledge of biological processes that are recognized series of events or molecular functions. The FrameNet is a lexical resource for general English, based on the theory of {\em frame semantics}
and supported by corpus evidence. There is a network of frames that are linguistically-oriented classifications of semantics. A semantic frame describes an event or a situation and the relevant participants and roles. For example, {\sf Progress} defines the situation that ``an {\sf Entity} changes from a {\sf Prior\_state} to a {\sf Post\_state} in a sequence leading to improvement''. 
A set of words (not only verbs) are grouped into a frame based on their semantic meaning.


The related work include the BioFramenet~\cite{Dolbey09}. It is an extension of the FrameNet with domain-specific frames. Its semantic frames are based on 5 protein transport classes from the Hunter Lab protein transport knowledge base. 
In~\cite{SKJ09}, the GENIA group made an effort to investigate the gap between its domain-specific corpus to FrameNet by associating 4 of GENIA event concepts with the FN semantic frames. 

\section{Method}

To collect the FN semantic frames (releases 1.3) that are mostly likely to be valuable for describing and processing text in molecular biology domain, we utilized biomedical verbs present by PASBio
project (releases 1.0).  
It represents 29 verbs (representing 34 predicates) chosen for their usage in the description of gene expression and related events. 
The assignment of a FN frame to a verb is based on the definition of PAS predicates and their arguments, FN frames and their FEs. A FN frame became a candidate frame for a verb, if the verb is a LU in the FN frame. We discarded  FN frames that either have completely different meanings to the sense of the verb, or are too specific for biological situations. If there were no entry for the verb, or all available frames are discarded, synonyms or words in its definition were used. 
We manually examined all PASBio predicates and their arguments, and FN frame candidates and their FEs, and made the assignments. 19 FN semantic frames are collected in this stage.

In the biological process ontology of GO (releases 1.2 in OBO format), events and molecular functions are named by using noun forms corresponding to verbs.
A biological process concept becomes a candidate that is semantically relevant to a FN frame, if the head of the concept name is the noun form of a verb that is a LU in the FN frame. The second step is to choose GO concepts from candidates that are the mappings to the FN frames. Here we define the mapping as ``{\it the GO concept can be seen as a subclass of the FN semantic frame in biomedicine domain}''. More general concept is more likely to be considered as the concept mapping to a FN frame than more specific concept, since a general concept subsumes all specific concepts. 
If there is part-whole relationship between two candidates, the part concept is less likely to be considered the concept mapping to a FN frame. The reason is that the whole concept includes the part concepts.
The GO mapping candidates were first collected automatically by evaluating concept names. Then they are returned as the mappings if a candidate is neither a subclass of other concepts nor a part of other concepts. The program is written in release 0.11 of perl modules for GO and OBO ontologies. Then we manually examined all mapping candidates and identified the final mappings. All the final mappings between the FN and GO are available on http://www.ida.liu.se/~hetan/FN2GO.

\section{Result and Discussion}

The resulting mappings between GO concepts to the FN frames can be explained by decomposing it into different cases. 

{\bf No mapping GO concept.} For four FN frames, {\sf Activity\_start}, {\sf Causation}, {\sf Interrupt\_process}, {\sf Forgoing},  there are no semantically relevant GO concepts. While their words 
are used  to describe biological events or situations, there are no defined GO concepts that represent the events or situations expressed by them.

{\bf 1 mapping GO concept.} The GO concept {\sf GO:0008283:cell proliferation} is associated with the FN frame {\sf Proliferation\_in\_number}. We noticed that in the four subclasses of  {\sf GO:0008283:cell proliferation}, {\it cyst formation} describes the formation of a cluster of germ-line cells derived from a single founder cell. 
While in FN {\it form} and {\it formation} are the LUs of the FN frame {\sf Creating}, the phrase {\it cyst formation} can be grouped into such a frame in the phenomenon of biology.
Secondly, the GO concept {\sf GO:0006412:translation} is associated with the FN frame {\sf Translating}. In its subclasses, the head of concepts name is {\it translation}.

{\bf Multiple mapping GO concepts.}
Two situations exist in the case: 
1) {\it  Multiple verbs.} There are two examples in this situation. Two GO concepts {\sf  GO:0007571:age-dependent general metabolic decline} and {\sf GO:0040007:growth} are associated with the FN frame {\sf Change\_position\_on\_a\_scale}.
The frame profiles words that indicate the change of an {\sf Item}'s position on a scale ({\sf Attribute}) from a starting point ({\sf Initial\_value}) to an end point ({\sf Final\_value})''. The words {\it decline} and {\it growth} describe biological process of opposite directions.  
Three GO concepts, {\sf GO:0018409:peptide or protein amino-terminal blocking}, {\sf GO:0018410:peptide or protein carboxyl-terminal blocking} and {\sf GO:0060468:prevention of polyspermy}  are associated with the FN frame {\sf Preventing}. They describe three very different kinds of biological process by using 
 words, {\it prevent} and {\it block}.  
The head of the concept names of the subclasses of {\sf GO:0018409} and {\sf GO:0018410}, including {\it carbamoylation} and {\it carboxylation}, are not LUs of the frame {\sf Preventing}. They describe specific chemical reactions. Only when these actions happen to N-terminal protein amino acid, preventing situation happens to the protein or peptide sequencing. This kind of domain-specific verbs cannot be grouped into any frames. 
2) {\it Multiple specific biological process.}
For the rest of the FN frames, there are a set of very specific mapping GO concepts. The frames include {\sf Removing}, {\sf Cause\_change}, {\sf Cause\_change\_of\_position\_on\_a\_scale}, {\sf Progress}, {\sf Creating} and {\sf Becoming\_aware}.  
For example, the GO concepts mapping to the frame {\sf Removing}, are located in the different branches of the GO, such as {\sf GO:0006915:apoptosis}, {\sf GO:0043412:macromolecule modification} and {\sf GO:0006259:DNA metabolic process}. No general concept subsumes the specific events or situations.  

{\bf GO concepts mapping to multiple FN frames.}
In this case, the first situation can be explained by using the example {\sf GO:0032502 developmental process}. 
Based on the definition, this concept can be the mapping to the frame {\sf Progress}.  However, many of its subclasses describe a biological progress not semantically relevant to {\sf Progress}. For example, the subclass {\sf GO:0010014:meristem initiation}, can be considered as the mapping to the frame {\it Process\_start}. On the other hand, the head of the concept names of some subclasses are not LUs of neither the frame {\sf Progress} nor any frames in FrameNet, but the concept describes the situation semantically relevant to {\sf Progress}, e.g. the concept {\sf GO:0043934:sporulation}.
The second situation refers to the concept {\sf GO:0019882:antigen processing and presentation}. The verb {\it present} is in the LUs of the frame {\sf  Cause\_to\_perceive}, and {\it process} is the word referring to the frame {\sf Process}. This kind of concepts cover a series of events and can not be simply mapped to one frame.

The result shows the gap between the classification of events or situation in GO and FrameNet. From the point of view of biology, GO may describe a situation that happens to different substances by different specific concepts, but it is defined as a semantic frame in the point of view of linguistics. A general GO biological process concept may cover different events that are described by different semantic frames, although they all are considered as a kind of the general biological process from the point of view of biology. For some events, although they appear in the context in molecular biology domain, the semantics are general.    
Secondly, the result illustrates the gap between the grouping of words in the domain and in general English. The granularity of groupings can be different from the point of view of linguistics and biology. It is also difficult to create semantic frames for some domain-specific words.

\end{document}